%% file: rahul_kamal_etal.tex
\documentclass[letterpaper, 10 pt, conference]{ieeeconf}  

\IEEEoverridecommandlockouts                              

\let\labelindent\relax
\usepackage{enumitem}
\usepackage{graphics} 
\usepackage{epsfig} 
\usepackage{mathptmx} 
\usepackage{times} 
\usepackage{amsmath} 
\usepackage{amssymb}  
\usepackage{booktabs}
\usepackage{graphicx}
\usepackage{tabularx}
\usepackage{multicol}
\usepackage{algorithm}
\usepackage{algorithmic}
\usepackage{multirow}
\usepackage{todonotes}%
\usepackage{lettrine}
\usepackage{subcaption}
\usepackage{wasysym}
\usepackage{bm}

\usepackage[linewidth=1pt]{mdframed}
\usepackage{setspace}

\newcommand{\+}[1]{\ensuremath{\mathbf{#1}}}

\graphicspath{{pictures/}}
\usepackage{cite}
\usepackage[bookmarks=false]{hyperref}
\usepackage{leftidx}

\definecolor{lightlightgrey}{rgb}{0.9,0.9,0.9}
\definecolor{Red}{rgb}{1,0,0}
\definecolor{Blue}{rgb}{0,0,1}
\definecolor{Green}{rgb}{0,1,0}
\definecolor{magenta}{rgb}{1,0,.6}
\definecolor{lightblue}{rgb}{0,.5,1}
\definecolor{lightpurple}{rgb}{.6,.4,1}
\definecolor{gold}{rgb}{.6,.5,0}
\definecolor{orange}{rgb}{1,0.4,0}
\definecolor{hotpink}{rgb}{1,0,0.5}
\definecolor{newcolor2}{rgb}{.5,.3,.5}
\definecolor{newcolor}{rgb}{0,.3,1}
\definecolor{newcolor3}{rgb}{1,1,1}
\definecolor{darkgreen1}{rgb}{0, .35, 0}
\definecolor{darkgreen}{rgb}{0, .6, 0}
\definecolor{darkred}{rgb}{.75,0,0}

\xdefinecolor{olive}{cmyk}{0.64,0,0.95,0.4}
\xdefinecolor{purpleish}{cmyk}{0.75,0.75,0,0}

\xdefinecolor{red}{rgb}{1,0,0}

\usepackage{float}
\floatstyle{boxed}
\newfloat{math}{thp}{lop}
\floatname{math}{Equation}

\makeatother

\begin{document}

\title{Energy Conscious Over-actuated Multi-Agent Payload Transport Robot: \\Simulations and Preliminary Physical Validation}
\author{Rahul Tallamraju$^{1,2}$, Pulkit Verma$^1$, Venkatesh Sripada$^3$, Shrey Agrawal$^1$ and Kamalakar Karlapalem$^1$
\thanks{\hspace{-1em}rahul.tallamraju@tuebingen.mpg.de,sripadav@oregonstate.edu, {pulkit.verma,shrey.agarwal}@research.iiit.ac.in,kamal@iiit.ac.in}
\thanks{\hspace{-1em}$^1$Agents and Applied Robotics Group, IIIT Hyderabad, India.}
\thanks{\hspace{-1em}$^2$Max Planck Institute for Intelligent Systems, T\"ubingen, Germany.}
\thanks{\hspace{-1em}$^3$Oregon State University, Corvallis, United States.}
}
\maketitle
\pagestyle{empty}
\begin{abstract}
\input{abstract}
\end{abstract}


\input{intro_new}

\input{related_work}

\input{model}

\input{poa}

\input{sim_experiments}

\input{conclusions}


\bibliographystyle{IEEEtran}
\bibliography{paper}

\end{document}

%% file: abstract.tex
In this work, we consider a multi-wheeled payload transport system. Each of the wheels can be selectively actuated. When they are not actuated, wheels are free moving and do not consume battery power. The payload transport system is modeled as an actuated multi-agent system, with each wheel-motor pair as an agent. Kinematic and dynamic models are developed to ensure that the payload transport system moves as desired. We design optimization formulations to decide on the number of wheels to be active and which of the wheels to be active so that the battery is conserved and the wear on the motors is reduced. Our multi-level control framework over the agents ensures that near-optimal number of agents is active for the payload transport system to function. Through simulation studies we show that our solution ensures energy efficient operation and increases the distance traveled by the payload transport system, for the same battery power. We have built the payload transport system and provide results for preliminary experimental validation.

%% file: intro_new.tex
\section{Introduction} \label{sec:intro}
The advent of electric vehicles and related emerging technologies \cite{todd2013creating} are the result of understanding an urgent need for  clean and safe energy for human or cargo transportation. Over-actuated systems have advantages over three or four-wheeled mobile robot systems in  applications like traversing uneven terrain or transporting heavy loads ~\cite{wilcox2007athlete,iagnemma2000mobile,turlapati2015stair}. However energy efficient, wear conscious and failure resilience properties of such multi-wheel systems have not been extensively studied in literature. {Our work aims to enhance efficiency and motivate the use of multi-agent control for transportation systems.}
In this paper, we introduce an over-actuated multi-agent payload transport robot (MAPTR) which uses multiple low-power motors (refer Table \ref{tab:1}) to achieve near energy efficient motor operation. 
The presented methodology actuates only a subset of all motors at near constant velocities based on the motor energy efficiency criterion. Our results provide evidence for such an energy conscious operation. 
The primary challenges in achieving energy conscious operation for an over-actuated robot can be categorized into (1) system kinematic and dynamic modeling, (2) controller design for energy efficient actuator control, and (3) real-time computation of control allocation.

\noindent We address these challenges as follows. \\
\textbf{1. Modelling the robot kinematics and dynamics.} 
Each wheel of the robot rotates about its associated motor shaft axis, which is its only degree of freedom. We therefore use  skid-steer kinematic and dynamic models (Sec. \ref{sec:3}, Sec.\ref{sec:5}) to manoeuvre the system along any trajectory.\\
\textbf{2. Energy and wear conscious operation.}
As the system moves from rest and achieves near constant velocity, the torque required to maintain a reference velocity can be distributed among a subset of motors. Therefore, we explore effects of deactivating motors while the robot is in motion. Each wheel in the robot is associated with a low-power, low-torque gearless motors (refer Table \ref{tab:1}) to enable activation (actuation) and deactivation. By using fewer motors at nearly constant velocity, the system ensures that each motor (or agent) can function about its energy efficient operating point (Sec. \ref{sec:52}), thus making the system conscious of its energy usage. Additionally, since only a subset of agents is active, we can distribute the effort of transporting the load among different active-inactive agent sets, thereby making the system wear-conscious (Sec. \ref{sec:54}). \\
\textbf{3. Multi-level decision-making controller.} The system is modeled as a multi-agent system where each wheel-motor pair is considered as an agent. The agent has control over its wheel velocity. A group of agents are controlled by a group controller which allocates activation and deactivation signals based on energy efficiency. Multiple group controllers communicate to an on-board central controller which provides high-level trajectory information for the robot.\\

We validate the proposed robot model and its energy conscious operation in simulation experiments. Further, we verify the feasibility of using multiple low-power gearless motors to transport payloads with a real over-actuated robot.\\
%
\begin{figure} [h]
	\centering
	\includegraphics[width = \columnwidth]{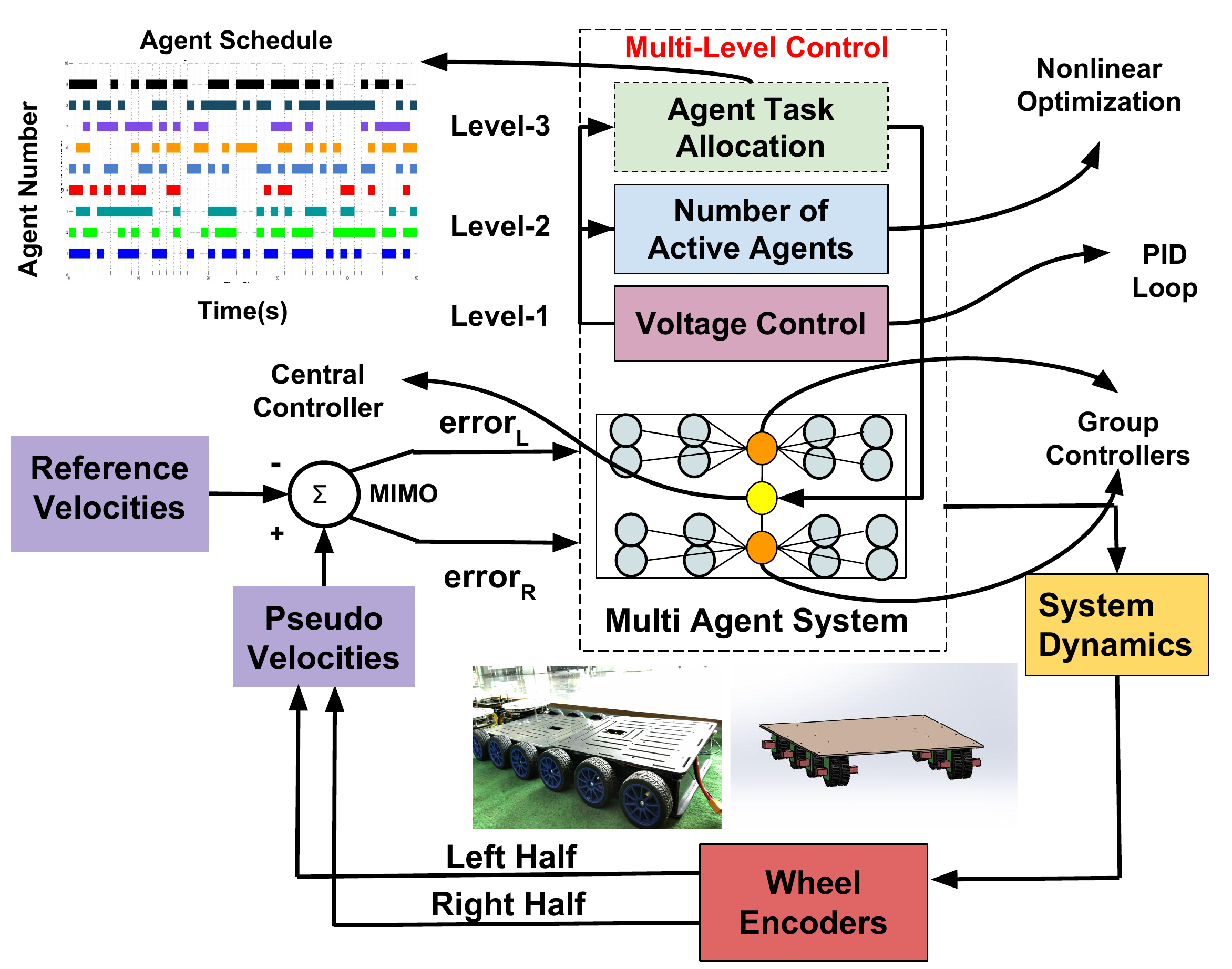}
	\caption{The three level control architecture with a snapshot of the real multi-wheel system.}
	\label{Fig:1}
\end{figure}
 \begin{figure*}
	\begin{subfigure}[t]{0.49\linewidth}
		\centering
		\includegraphics[ width = \columnwidth ]{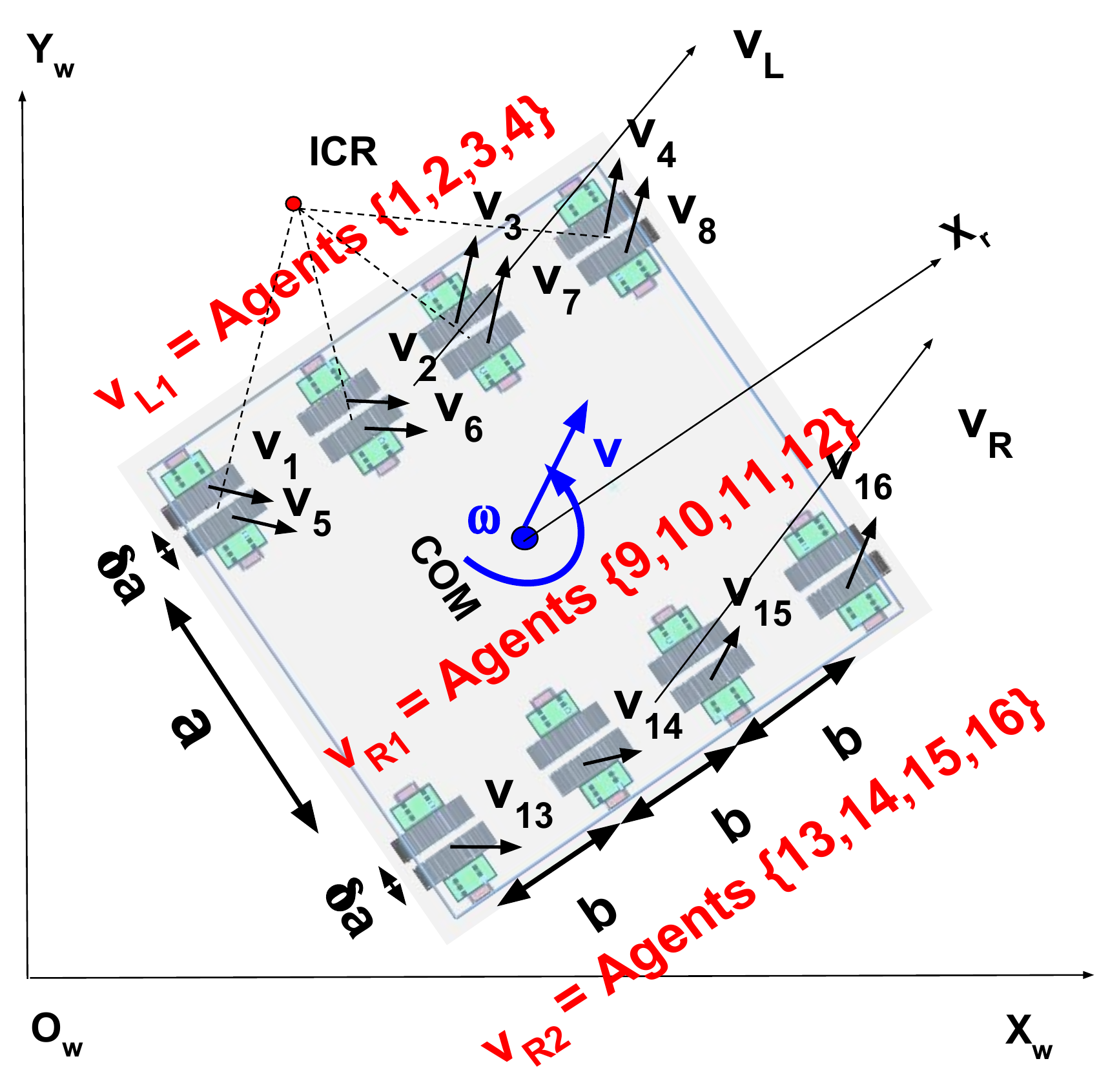}
		\caption{Kinematic parameters}
		\label{VModel}
	\end{subfigure}
	\begin{subfigure}[t]{0.49\linewidth}
		\centering
		\includegraphics[width = \columnwidth ]{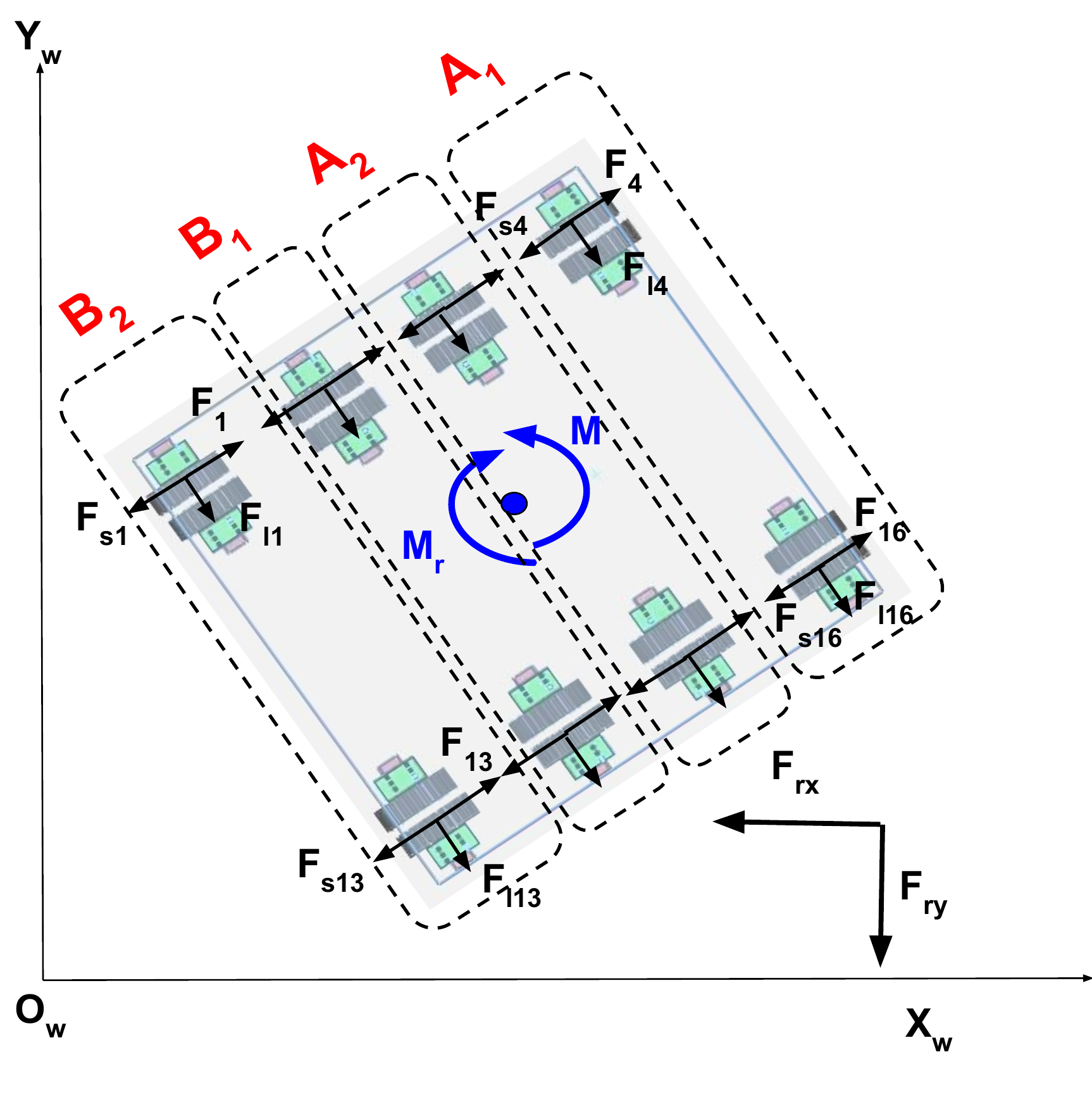}
		\caption{Forces acting on system}
		\label{FModel}
	\end{subfigure} 
	\vspace{0.3cm}	
	\caption{Kinematic and Dynamic Parameters of an illustration of the system}
	\label{ForceV}       
\end{figure*}

%% file: related_work.tex
\section{Related Work} \label{sec:related_work}
 Advantages of distributed electric control with redundant actuators for future aerial systems are discussed in ~\cite{holden2016fast,moore2003personal,gigante2017volocopter}. 
{ To the best of our knowledge not much of work has been carried out in identifying energy conscious and wear minimizing control allocation for over-actuated autonomous ground vehicles. Therefore, we review literature related to different aspects of our work. }
Energy efficient control allocation for electric vehicles was studied to track longitudinal trajectories in \cite{chen2012fast,chen2014design}. In that work, actuation and re-generative breaking controls are allocated to the motors to achieve energy efficient operation using a piecewise linear motor efficiency model.  
{ In \cite{tokekar2011energy} energy optimal acceleration-deceleration profiles are identified for unmanned ground vehicles through a novel actuator calibration and modelling procedure. These profiles are generated offline for a given battery lifetime and trajectory of a four wheeled robot.}
{ Fault tolerant control allocation for over-actuated systems was explored in \cite{khelassi2010reconfigurable}. Control signals are dynamically allocated to actuators through a control effectiveness matrix and a numerical optimization formulation. The allocation was however only applied to LTI systems.}
Wheel-motor pairs are modeled as independent agents \cite{dziomin2013multi} to control a wheeled robot along a circular trajectory using multi-agent reinforcement learning. However not more than four driving modules are used and also motor wear reduction is not explored.  
{In contrast to the aforementioned literature, our work ensures that a collaborative effort by a large group of motors achieves energy conscious operation. This makes it conducive to model our system as a networked multi-agent system \cite{ota2006multi,parker2008multiple,chevaleyre2005multiagent,durfee1994distributed}. Moreover, agents in the system are selectively actuated in real-time based on system energy efficiency, system trajectory and work done criteria. The two key steps which enable such an agent actuation scheme are, 
(1) system dynamics constrained, energy efficient actuator control-allocation, (2) control-allocation constrained, wear minimizing actuation and deactivation agent schedules. } 

%% file: model.tex
\section{Robot System Model}  \label{sec:3}
The MAPTR is modeled as a multi-agent system. Each wheel-motor pair is modeled as an \textbf{agent} and a set of agents forms a \textbf{group}. Agents in a group are associated with a controller, limited battery power, wired connection to the group controller and local sensing in the form of wheel encoders. In this paper, agents are assigned to control groups, \textit{ left half group} ($L$) and \textit{right half group} ($R$)\footnote{These control groups can be further divided into sub-groups in the case of a cascaded multi-robot system, where each robot controls its own motors}, based on their spatial positions with respect to the central controller.  The group controllers are further connected to a system level central controller. It should be noted that in the MAPTR the wheels move freely when the motor is powered off. The free movement is hindered only by wheel-ground contact friction and low viscous friction in the gearless motor shaft. The full control architecture of the MAPTR is as shown in Fig. \ref{Fig:1}. Control at various levels is summarized below. \textbf{Level-1 (agent decision)} helps in maintaining a desired wheel velocity (Sec. \ref{sec:51}). \textbf{Level-2 (group and system decision)} determines the required number of active agents for near energy efficient actuator operation (Sec. \ref{sec:52}). \textbf{Level-3 (group decision)} identifies an agent activation-deactivation schedule to ensure uniform usage of all the agents. The controller takes decisions based on the following inputs (a) active agent velocity (level-1), (b) active agent efficiency (level-2) and (c)  number of active agents (level-3). For concept demonstration, we consider \textbf{(a)} identical agents, \textbf{(b)} agents belonging to the same group are coupled along their motor shaft axis (thereby forming a multi-motor drive), \textbf{(c)} flat terrain, \textbf{(d)} no longitudinal wheel slip. 
Ideally, with small in-hub motors, columns of actuated wheels can be stacked on either side of the system. A sixteen wheel compact configuration with two columns of closely spaced wheels on either side of the system center of mass (COM) is chosen for concept demonstration. We derived kinematic, dynamic and electrical models below.

\subsection{Kinematic Model}
The MAPTR uses skid-steering to manoeuvre, therefore, lateral velocity of the wheels is non-zero. Figure \ref{VModel} shows spacing between agents for $16$ wheel model.  The system has a linear velocity, {\small$\+{v} = [v_x, v_y, 0 ] ^T$}\normalsize 
and angular velocity {\small$\+{\omega} = [ 0, 0, \omega ]^ T$}\normalsize.  In Fig. \ref{VModel} subscripts $[L,R]$ represent the left and right control groups and $[A,B]$ represent the front and back wheel sets, the wheel velocities are  related  as follows. 

A design constraint {\small$\delta a << \frac{a}{2}$}\normalsize is considered then {\small$v_L = v_{L1} = v_{L2}$ and $v_R = v_{R1} = v_{R2}$. \small  $ \{v_{L} = v_{jx} \mid j \in {1-8}\}$,  $\{v_{R} = v_{jx} \mid j \in {9-16}\}$, $\{v_{A1} = v_{jy} \mid j \in {4,8,12,16}\}$, $\{v_{A2} = v_{jy} \mid j \in {3,7,11,15}\}$, $\{v_{B1} = v_{jy} \mid j \in {2,6,10,14}\}$, $\{v_{B2} = v_{jy}\mid j \in {1,5,9,13}\}$}\normalsize.  
\begin{eqnarray} \label{wheelV3}
 \begin{bmatrix}
v_{L} &  v_{R} & v_{A1} & v_{A2} & v_{B1} & v_{B2}
\end{bmatrix}^T = \boldsymbol{\Lambda}_{6 \times 2} \begin{bmatrix}
v_x &  \omega
\end{bmatrix}^T \\
where, \bm{\Lambda_{6 \times 2}} = \begin{bmatrix} 1  & 1 & 0 & 0 & 0 & 0  \\ -\frac{a}{2} & \frac{a}{2} & \frac{3b}{2} & \frac{b}{2} &  -\frac{b}{2} & -\frac{3b}{2}  \end{bmatrix}^T\nonumber
 \end{eqnarray}
System pose in generalized coordinates is \small $\+{q} = [X, Y,  \theta, \phi_{L},\phi_{R}] ^T$\normalsize, where, {\small$[X, Y,\theta]$} is the position and the orientation of the of the system in the world frame and {\small $[\phi_{L}, \phi_{R}]$}  are the angular wheel orientations about their rotation axis. From equation (\ref{wheelV3}), the pseudo velocities \small $\+{\nu} = \begin{bmatrix}
v_x & \omega \end{bmatrix}^T = r\boldsymbol{\Lambda}_{2\times 2} \begin{bmatrix} \dot{\phi_L} & \dot{\phi_R} 
\end{bmatrix}^T $\normalsize, where $r$ is the wheel radius. The generalized and pseudo velocities ($\dot{\+{q}}, \+\nu$) are related through the kinematic constraint matrix $\+S(\+q)$, as given in eqn. (\ref{kinematics}), which spans the null space of the non-holonomic constraint matrix ($v_y = 0$ or $-\dot{X} sin{\theta}+ \dot{Y} cos\theta = 0$). 
\begin{eqnarray} \label{kinematics}
 \dot{\+{q}} &=& \+S(\+{q}) \+\nu, \\
  \+S(\+{q}) &=& \begin{bmatrix}
  cos(\theta) & sin(\theta) & 0 & 1 & 1 \\
  		0		  & 			0		 & 1  & -\frac{a}{2} &\frac{a}{2} \end{bmatrix}^T \nonumber   
\end{eqnarray} 

\subsection{Dynamic Model} \label{MADynamics}
Referring to Fig. \ref{FModel}, and solving the Lagrange-Euler equation and, further eliminating Lagrange multipliers for non-holonomic constraints, the system dynamic model
is given by,
\small \begin{eqnarray} 
 \+{S}^T(q)\+M(q)\ddot{\+q} + \+S^T(q)\+R(\dot{\+q}) &=& \+S^T(\+q)\+B(\+q) \boldsymbol{\tau}  \label{fdynamics}
\end{eqnarray},   \normalsize
where, {\small $\+M(q)$= diag($m$, $m$, $I_{zz}$, $N_LI_w$, \ $N_RI_w$)}. Here, $m$ is the mass of the system, $I_{zz}, I_w$ are mass moment of inertia of the chassis about the z-axis and the wheel about its rotation axis, $\ddot{\+q}$ is the acceleration in generalized coordinates, {\small $\+R(\dot{\+q}) = [F_{rx}, F_{ry}, M_{r}]$} is a generalized frictional force vector \cite{kozlowski2004modeling} along longitudinal, lateral and yaw axes.  
\small $F_{lj} =\mu_{l} m g \frac{2}{\pi}{tan^{-1}(k_s v_{yj})}$\normalsize  is the lateral friction on wheel, \small $F_{sj} =\mu_{s} m g \frac{2}{\pi}tan^{-1}(k_s v_{xj})$\normalsize is the longitudinal friction on wheel. The viscous friction is modeled as part of DC motor dynamics (Sec. \ref{sec:41}). $\mu_s, \mu_l$ are the coefficients of rolling friction and lateral sliding friction for the wheels with ground, $g$ is the  gravitational acceleration and  $k_s >> 1$ is a constant.  ${\+B(\+q)}$ is the input transformation matrix and the total system torque applied by the motors given by, {\small $ \boldsymbol{\tau} =
\begin{bmatrix}
\sum_{i \in L} \tau_i, \sum_{i \in R} \tau_i
\end{bmatrix}^T = [\tau_L,\tau_R]^T$}. 
Final state-space dynamics update equation is derived by differentiating equation (\ref{kinematics}) ($\ddot{\+{q}} = {\dot{\+S}(\+q)}\+{\nu} + {\+S(\+q)}\dot{\+{\nu}}$) and substituting it in equation (\ref{fdynamics}). 
\begin{equation} \label{dynamics_final}
\bm{\dot{\nu}} = \begin{bmatrix}
\dot{v}_x  \\ \dot{\omega}
\end{bmatrix} = \begin{bmatrix}
\frac{\tau_L + \tau_R - r(F_{rx}(\bm{\dot{\+q}}) cos(\theta)+F_{ry}(\bm{\dot{\+q}}) sin(\theta))}{r(m+16I_w)} \\
\\
\frac{-a \tau_L + a \tau_R - 2r M_r (\bm{\dot{\+q}})}{2r(I+\frac{8I_wa^2}{2})}
\end{bmatrix}
\end{equation}\normalsize
%
\subsection{Integrating Motor and Battery Models } \label{sec:41}
\vspace{-1.2em}
The MAPTR has a motor associated with each wheel and a battery associated with each group controller, it is therefore pivotal to model their characteristics.
\subsubsection{Motor Model} Using the first order DC motor model, actuator torques in the system are given by,

\begin{eqnarray} \label{motor1}
\tau_i &=& N_{a,i} (K_T I_{a,i} - K_T I_0 - b_{damp}\dot{\phi}_i), \ I_{a,i} = \frac{V_i - K_e \dot{\phi}_i}{\Omega} \nonumber \\
\ i &\in& [L,R], \ \tau_{total} = \tau_L + \tau_R, \ N_a = N_{a,L} + N_{a,R}
\end{eqnarray}
where, $V_L, V_R$ are left and right group motor supply voltages, $\Omega$ is the armature resistance, $K_e$ is the  back emf constant, $\dot{\phi}_{i}$ is the motor shaft angular velocity, $K_T$ is the motor torque constant, $I_0$ is the no-load current and $b_{damp}$ is the viscous friction coefficient, $N_{L}, N_{R}$  are total number of motors in the left and right groups, and $N_{a,L}, N_{a,R}$ are the total number of active motors in the left and right groups. The motor integrated dynamic model is obtained by substituting eqn. (\ref{motor1}) in (\ref{dynamics_final}) as shown below.
\begin{eqnarray} \label{integrated_dynamics}
&\dot{\+{\nu}} = [\dot{v}_x \ \dot{\omega}] = \\ & \begin{bmatrix}
\frac{\tau_{total}- (N-N_a) b_{damp} - r(F_{rx}(\bm{\dot{q}}) cos(\theta)+F_{ry}(\bm{\dot{q}}) sin(\theta))}{r(m+16I_w)} \\
\\
\frac{a(- \tau_L +  \tau_R +b_{damp}((N_L-N_{(a,L)}) - (N_R-N_{(a,R)}) )) -  2r M_r (\bm{\dot{q}})}{2r(I+\frac{8 I_wa^2}{2})} 
\end{bmatrix} \nonumber
\end{eqnarray}

where, $N = N_L+N_R$ and the additional damping terms $(N_i - N_{a,i})b_{damp}$ are added because of the viscous rotational friction that exists due to the inactive motors in the system.

\subsubsection{Battery Model} The multi-wheel system will be operationally limited by its battery life. The battery voltage deterioration with discharge is experimentally determined.  We find the statistically best fit to the data as a two term exponential model $V_B=he^{wd_B}+ye^{zd_B}$, where, ${V_B}$ is the supplied battery voltage, $d_B$ is the battery discharge and $h,w,y,z$ are coefficients determined from the fit.  The results (Sec. \ref{sec:6}) showcase the effects of reducing battery voltage with and without efficiency optimization (Sec. \ref{sec:52}). 

%% file: poa.tex
\section{Multi-Level Control System} \label{sec:5}
The multi-level control system is as shown in Fig.\ref{Fig:1}. The MAPTR ensures that its agents operate efficiently. It does so by regulating the number of active agents, thereby ensuring to get greater work done with fewer active agents. Consequently, some agents in the system get a break from activity. The duration of the break is based on (i) motor models, (ii) battery consumption and (iii) system trajectory. 

\subsection{Level-1: Agent Velocity Control}  \label{sec:51}
Control at an agent level is achieved using proportional-integral-derivative (PID) control \cite{aastrom1995pid}. The central controller specifies reference wheel velocities for the agents in the left $\dot{\phi}_{ref,L}$ and right groups $\dot{\phi}_{ref,R}$.  At each discrete time instant $k$, an agent \footnote{$\dot{\phi}_i[k], \ i \in [L,R]$   velocities of an agent in a group are equal due to identical agent assumption} applies a control voltage 
$V_i[k]$.
\begin{equation} \label{PID}
V_i[k] = K_P e_i[k] + K_I E_i[k] + K_D \Delta e_i[k], \ i \in [L,R]~,
\end{equation}
where, $e_i[k] = \dot{\phi}_i[k] - \dot{\phi}_{ref,i}$, $E_i[k] = E_i[k-1] + e_i[k] $ and $\Delta e_i[k] =  e_i[k] - e_i[k-1]$. $K_P, K_I, K_D$ are gains corresponding to proportional, integral and derivative components of the PID, $E_i[k]$ is the cumulative error and $\Delta e_i[k]$ is the change in error. By tuning control gains, the PID achieves desired performance irrespective of the underlying MAPTR model. 

\begin{figure}[h]
	\centering
		\includegraphics[width = 9.2cm,height = 3.4cm]{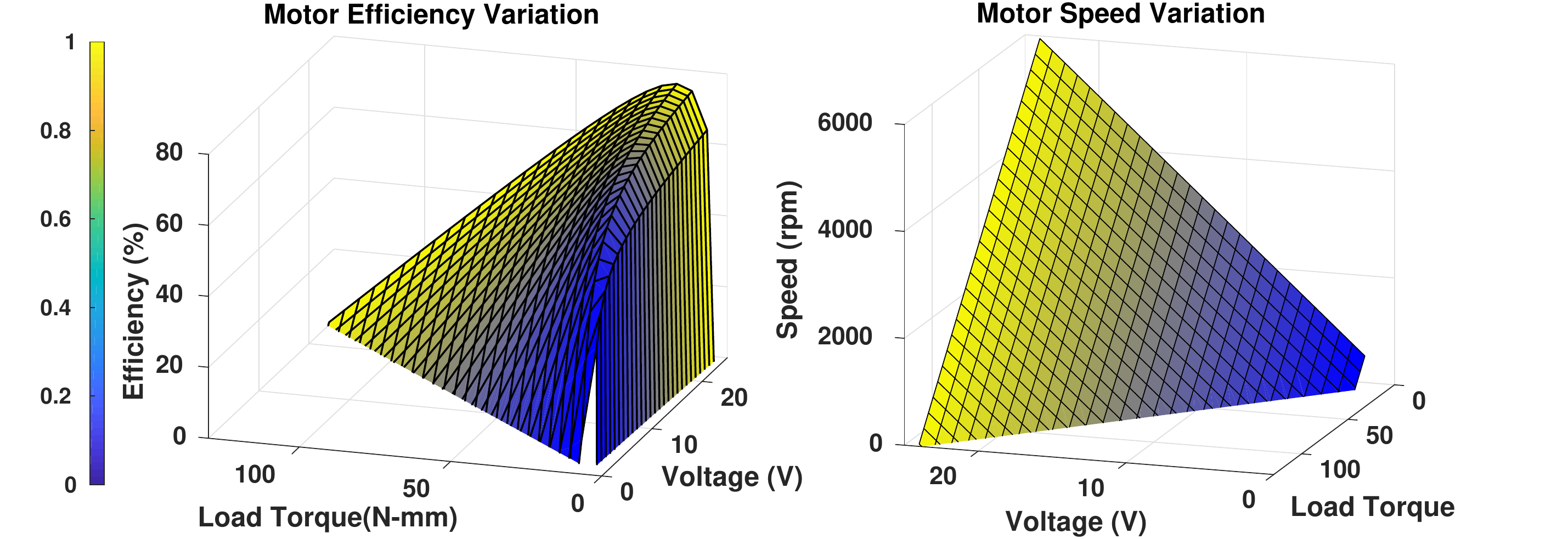} 
		\captionof{figure}{Efficiency v/s Torque and Speed v/s Torque}
		\label{motorTorque}
\end{figure}
\begin{table*}[]
	\centering
	\vspace{0.5cm}
	\caption{Mechanical and Electrical Parameters of the Multi-Agent System}
	\scriptsize
	\label{tab:1}
	\begin{minipage}{.3\textwidth}
		\begin{tabular}{@{}ccc@{}}
			Parameter        & Value & Units \\ \hline
			Number of Motors ($N$) & 4-128 &  -     \\
			Chassis Weight     & 0.2  & kg    \\
			Wheel Weight($m_w$)     & 0.05  & kg    \\
			Payload Weight ($m$)   & 10-60 & kg    \\
			a                & 0.2   & m     \\
			b                & 0.08  & m     \\
			Wheel radius($r$)     & 0.035 & m     \\
			Rolling Friction ($\mu_s$) & 0.01  &    -   \\
			Sliding Friction ($\mu_l$)& 0.1   &     - 
		\end{tabular}
	\end{minipage}
	\begin{tabular}{@{}cccc@{}}
		Parameter of Motor            & Type-I    			      			   & Type-II    		& Units     \\ \hline
		Rated Voltage ($V$)           & $24$                                     & $6$                  & $V$       \\
		No-load Current ($I_0$)       & $50$                                     & $250$                & $mA$      \\
		No-Load Speed ($\phi_0$)      & $5930$                                 & $5500$               & $rpm$     \\
		Stall Torque ($\tau_s$)       & $130$                                  & $17.6$               & $mN~m$     \\
		Armature Resistance ($R$)     & $7.03$                                 & $2.4$                & $\Omega$    \\
		Viscous Friction Coeff. ($b$) & $6 \times 10^{-7}$         & $2.2 \times 10^{-7}$ & $Nms$     \\
		Torque Constant ($K_T$)       & $38.2 \times 10^{-3}$  & $7 \times 10^{-3}$   & $Nm/A$    \\
		Back-emf Constant (K\_e)      & $38.4 \times 10^{-3}$   & $7 \times 10^{-3}$   & $Vs/rad$ \\
		Motor Weight 	              & $0.21$                       & $0.11$                  & $kg$ 
	\end{tabular}
	\normalsize
\end{table*}

\begin{figure*}
\begin{subfigure} {0.49\linewidth}
\centering
  \includegraphics[width=\columnwidth]{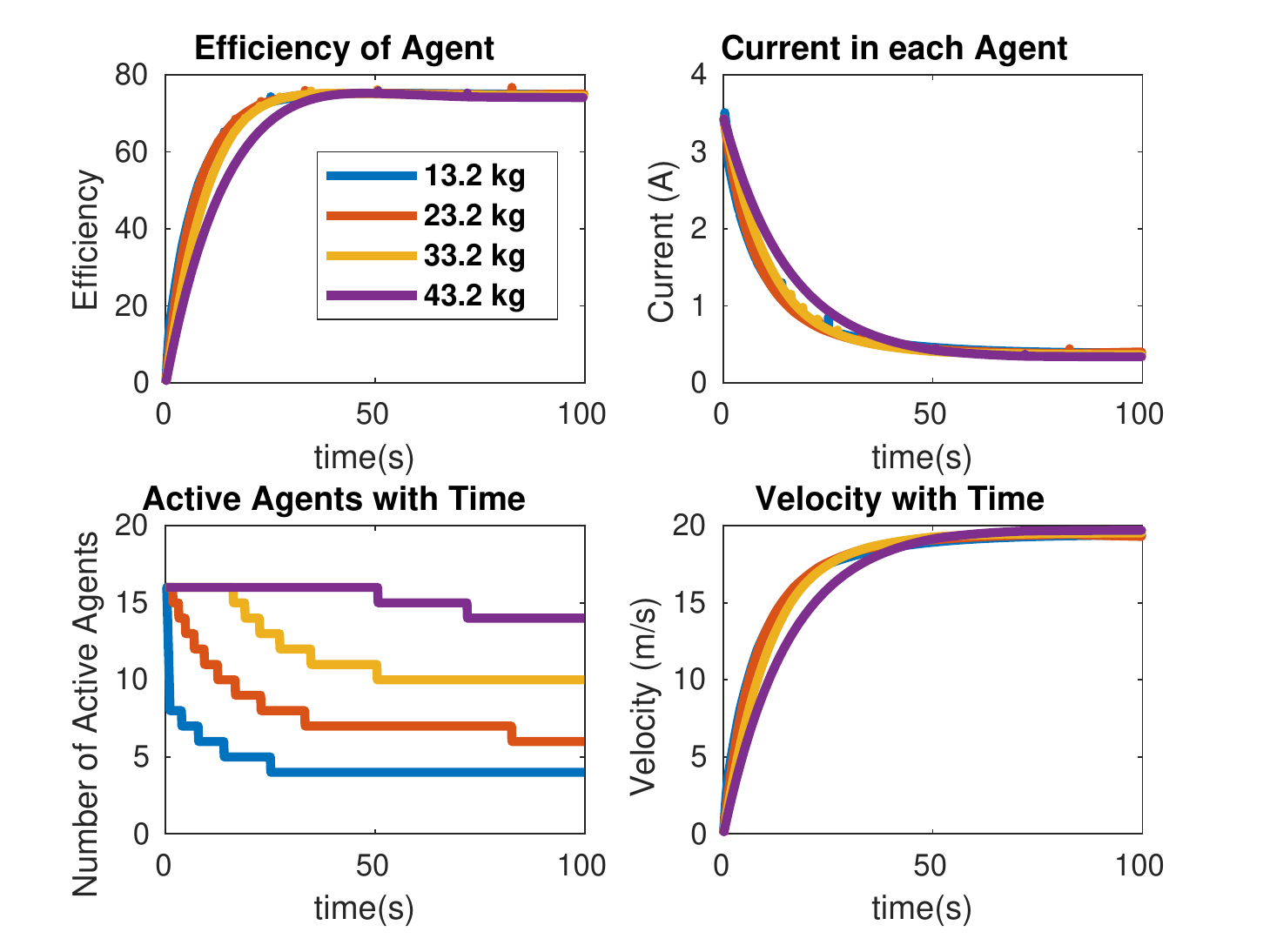}
  \captionof{figure}{ Level-2 : Different gross weights}
  \label{eff_result_1}
\end{subfigure}~ 
\begin{subfigure} {0.49\linewidth}
\centering
  \includegraphics[width=\columnwidth]{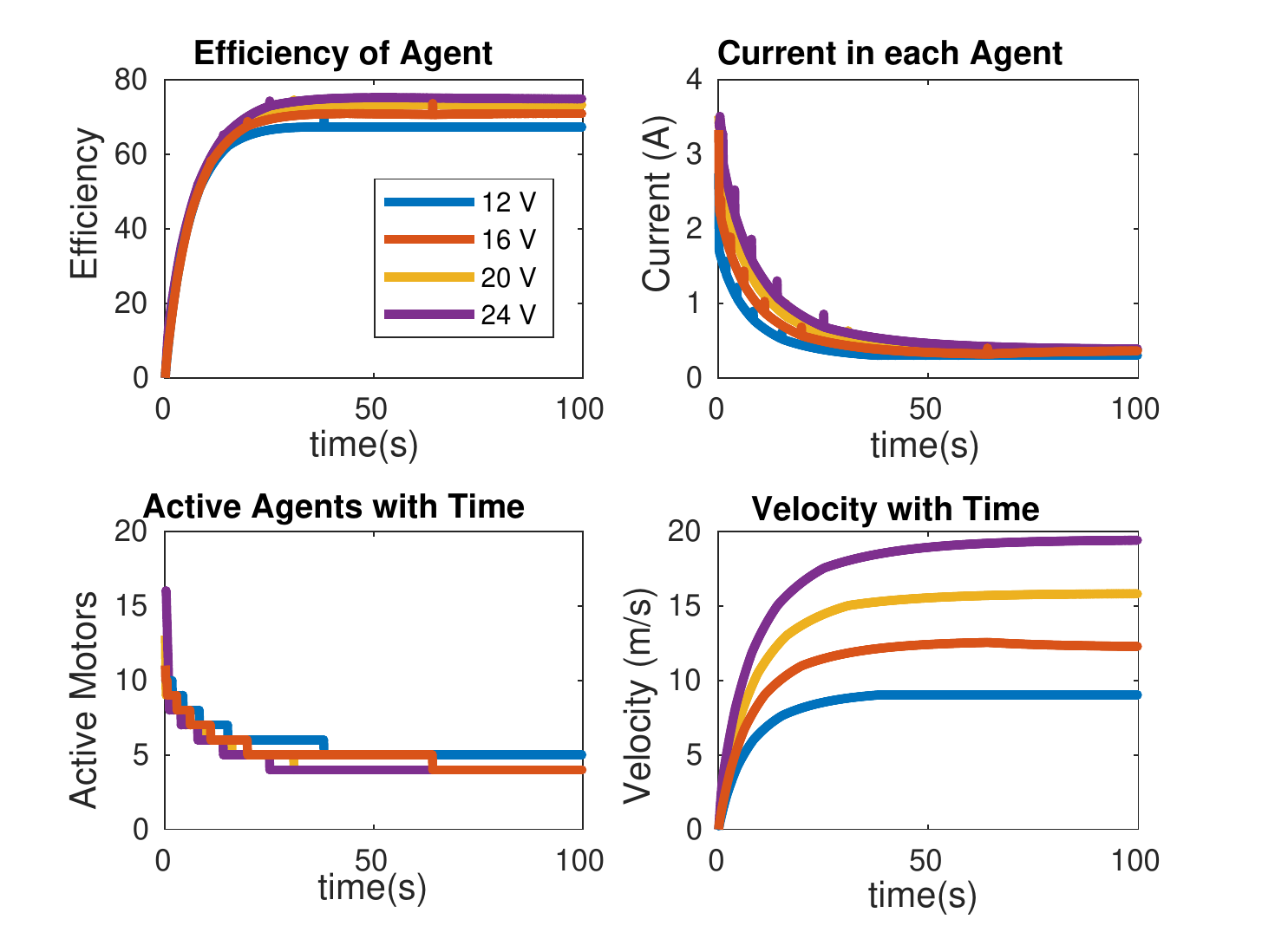}
  \captionof{figure}{ Level-2: Different input voltages}
   \label{eff_result_2}
\end{subfigure}
\vspace{0.5cm}
\caption{The above plots showcase the results of energy conscious operation for a 16-agent system. (a) Notice that numerically optimizing  (\ref{Efficiency}), for different payload weights, ensures that each agent (or motor) operates near its maximum efficiency ($\eta \approxeq 80 $\% from Fig. \ref{motorTorque}). (b) Notice that the agent achieves near maximum efficiency for different input voltages. The resulting efficiencies approximately match the characteristic curve in Fig. \ref{motorTorque}. }
\end{figure*}

\subsection{Level-2: Energy Conscious Operation} \label{sec:52}
\noindent  Surface plots in Fig. \ref{motorTorque} are characteristic curves of a typical permanant magnet DC motor (PMDC). The  curves are plotted using equation (\ref{motor1}) for, a single $24~V,~15~W$ motor with a stall torque of $130~Nmm$ and a no-load speed of $5930$ rpm. Efficiency ($\eta$) of a motor in the $n^{th}$ agent is calculated as $ \frac{\tau_n \dot{\phi}_{n}}{V_n I_{a,n}} $, where $\tau_n$  is the torque of $n^{th}$ agent, $\dot{\phi}_n$ is the wheel velocity, $V_n$ is the applied voltage and $I_{a,n}$ is the current drawn. From Fig. \ref{motorTorque}, we note that maximum motor efficiency lies  around $10-20 \%$ of stall torques at different voltages. Furthermore, we observe a steep drop in efficiency profile from maximum efficiency point to near zero efficiency at low load torques, due to relatively high dissipative losses. Efficiency of each motor in the MAPTR varies as a function of system load, number of active agents and dissipative losses.  

Voltage control (Sec. \ref{sec:51})  applies a voltage $V_i$ corresponding to the error with respect to reference velocity. To maximize  active agent efficiency by regulating the number of active agents, we formulate a non-linear optimization problem. Each group controller independently executes the optimization. The objective of the optimization is to maximize motor efficiency. It is constrained by the number of agents the group controller can actuate and the system dynamics. Since the dynamics depends on the torque supplied by both the agent groups, information on the number of active agents in neighboring group is communicated by the central controller. The optimization is defined as follows.
\begin{eqnarray}\label{Efficiency}
\begin{aligned}
\text{Maximize  } &  \sum_{h = 0}^H \frac{\tau_{i}[k+h] \dot{\phi}[k+h]}{V_i[k+h]I_{a}[k+h]}   \\
\text{s.t. }  & 0 \leq  N_{a,i} [k]  \leq N_i, i \in [L,R]  \\
& \dot{\phi}_{i}[k] = f(\tau_{a,L}[k-1],\tau_{R}[k-1],N_{a,L}[k],N_{a,R}[k]), \\
&I_{a,i}[k] = \frac{V_i[k] - K_e \dot{\phi_i}[k]}{\Omega_i},  \\
&\tau_i[k] =  K_T(I_{a,i}[k] - I_0) - b_{damp} \dot{\phi}_i[k]. 
\end{aligned}
\end{eqnarray}

%
Here, $N_{a,i}$ is the number of active agents in the left or right groups, {$f$ represents zero-order hold discretized dynamics of the system (\ref{integrated_dynamics}).}
The objective of optimization is to maximize agent efficiency (with identical agent assumption), by regulating number of active agents $N_{a,i}$. The cost is accumulated for a horizon $H$ because of instantaneous rise in current with agent deactivation, causing instantaneous decrease in agent efficiency. The advantage of agent deactivation can only be observed after a few future time steps. 
Energy conscious system operation ensures that the number of active agents required for efficient operation, converges to a constant value, for a smooth trajectory.  Since, the number of active agents ($N_{a,i}$) is less than the total number of agents in a group ($N_i$), number of allowed agent failures for a given payload weight is $(N_i - N_{a,i}), i \in [L,R]$ agents. Therefore the system remains operational despite some agent failures. {The optimization is feasible if the torque supplied by the total number of agents is greater than the system load torque. Furthermore, since the optimization is constrained by non-linear system dynamics, the problem being solved is non-convex. We resolve this by numerically evaluating the optimization using sequential quadratic programming \cite{boggs1995sequential}, where the given problem is locally approximated as a quadratic program. The approximation is refined over multiple iterations to achieve sub-optimal solutions. Therefore, for real-time control, the solution to (\ref{Efficiency}) is locally optimal albeit fast to compute. }

\subsection{Level-3: Online Task Allocation} \label{sec:54}
The MAPTR can switch to different configurations of active agents to uniformly share the work load. Agent task allocation decisions are taken independently by the group controllers. A configuration of active agents is enforced to be operational for a predefined period of time $T_{ON}$ (based on motor thermal model). The distance traveled by agents might vary for a non-smooth trajectory due to changes in number of active agents. Therefore, we present a constrained linear integer program to ensure that the distance traveled by each agent is approximately equal.  Optimization is evaluated every $T_{ON} (\approx 100)$ seconds or whenever the number of active agents ($N_{a,i}$) changes. As configuration changes are not rapid, the level-1 and level-2 operations remain unaffected. 
\begin{equation} \label{online}
\begin{aligned}
\text{Minimize \ } & (\+D[k] - \+D_{ref})^T \+x_i[k] \\
\text{s.t. \ } & \sum_{j=0}^{N_i} \+x_i[k](j) = N_{a,i}[k] \\
& 0 \leq \+x_i[k](j) \leq 1, \ \+x_i[k](j) \in \mathbb{Z}, \ \forall \	j \in [1,N_i]   \\	
where \ & i \in [L,R], \   
\end{aligned}
\end{equation}
In the above equation $\+D$ is the vector of  distances traveled by the agents. The values of $\+D$ are normalized to lie in $[0,1]$. $\+D_{ref}$ is a vector whose elements have a value $\frac{1}{N_i}$. The objective of the optimization evaluates the fairness of agent usage based on the distance travelled by each agent. 
This cost is evaluated as a weighted addition over a binary valued vector $\+x_i[k]$, which represents activity (1) or inactivity (0) of agents at instant $k$.  The first constraint imposes an equality over the number of active motors ($N_{a,i}[k]$), which was determined by the level-3 controller. The second constraint imposes binary values for the vector $\+x_i[k]$

%% file: sim_experiments.tex
\section{Results} \label{sec:6}
\noindent We showcase results obtained by
 (a) optimizing the straight line system efficiency for both constant and varied voltages for different payloads  (Sec. \ref{sec:52}), (b) computing optimal online task allocation (Sec. \ref{sec:54}), and, (c) testing the feasibility of using a collection of low-power motors in a real system. The various mechanical and electrical parameters (Motors Types-'I', 'II') considered for the multi-wheel system are as shown in Table \ref{tab:1}.
\\

\begin{table}[t]
\centering
\caption{\small Straight-Line power saving for a 16 agent system (distance of $36 km \ (approx.)$).  (EC-Energy Conscious Operation, NO-No optimization)}
\scriptsize
\label{tab:pow}
\begin{tabular}{@{}ccccccc@{}}
\toprule
\multirow{2}{*}{\begin{tabular}[c]{@{}c@{}}Gross\\  Weight (kg)\end{tabular}} & \multicolumn{2}{c}{Energy (kJ)} & \multirow{2}{*}{\begin{tabular}[c]{@{}c@{}}Power \\ Saving(\%)\end{tabular}} & \multirow{2}{*}{\begin{tabular}[c]{@{}c@{}}Active \\ Agents\end{tabular}} & \multicolumn{2}{c}{Mileage (m/J)} \\ \cmidrule(lr){2-3} \cmidrule(l){6-7} 
                                                                        & NO             & EC          			&                                         &                                        & NO              & EC              \\ \midrule
13.2                                                                & 92.33          & 63.17          & 31.58                                      &3                                      & 0.4098          & 0.5347            \\
18.2                                                                & 111.69        & 87.46         & 21.70                                       & 5                                     & 0.3359           &0.3970           \\
23.2                                                               & 131.05        & 108.94       & 16.87                                       & 6                                     & 0.2838           &0.3171           \\
28.2                                                                 & 150.41        & 132.81        & 11.7                                        & 8                                     & 0.2452           &0.2633           \\
33.2                                                                 & 169.77        & 156.58      	& 7.77                                          & 10                                  & 0.2153           & 0.2249           \\
38.2                                                                 & 189.13        & 180.35        & 4.64                                         & 12                                  & 0.1916           &0.1962          \\ \bottomrule
\end{tabular}
\normalsize
\end{table}

\textbf{Energy Conscious Operation:} To solve the numerical optimization of equations (\ref{Efficiency})  in real-time, we employ the sequential quadratic program (SQP) \cite{nocedal2006sequential}, with relaxed integer constraints. Since SQP solutions in non-linear optimization are sensitive to an initial solution guess, as a pre-processing step, we simulate and approximate the required number of  active agents that achieve efficient operation at different loads. A horizon of $10$ time steps results in  a stable number of active agents, with optimization running at $1Hz$. The optimization results for using Type-I motors with different gross system weights (payload + chassis weight) is presented in Fig. \ref{eff_result_1}. Through active agent regulation, the efficiency of each agent converges to nearly the same maximum value, despite increase in payload weight. Also, the number of active agents required, increases with increase in payload weight. Conversely, with increase in payload weight, the number of agents available in case of operational failures, decreases. If there are more than $N_i-N_{a,i}, i \in [L,R]$ failed agents, the system can no longer operate. The optimal number of active agents can also vary with a change in applied voltage to the motors. Effect of changing input voltage, for a $10$ kg payload is as shown in Fig. \ref{eff_result_2}. Table \ref{tab:pow} summarizes results of optimization for different gross system weights with the sixteen wheel system, moving along a straight trajectory for a duration of $30$ min ($36$ km). A significant power advantage is observed at lower payloads weights, which reduces as the payload weight nears the system rated weight. 
 
\begin{figure} 
	\centering
	\includegraphics[width=\columnwidth]{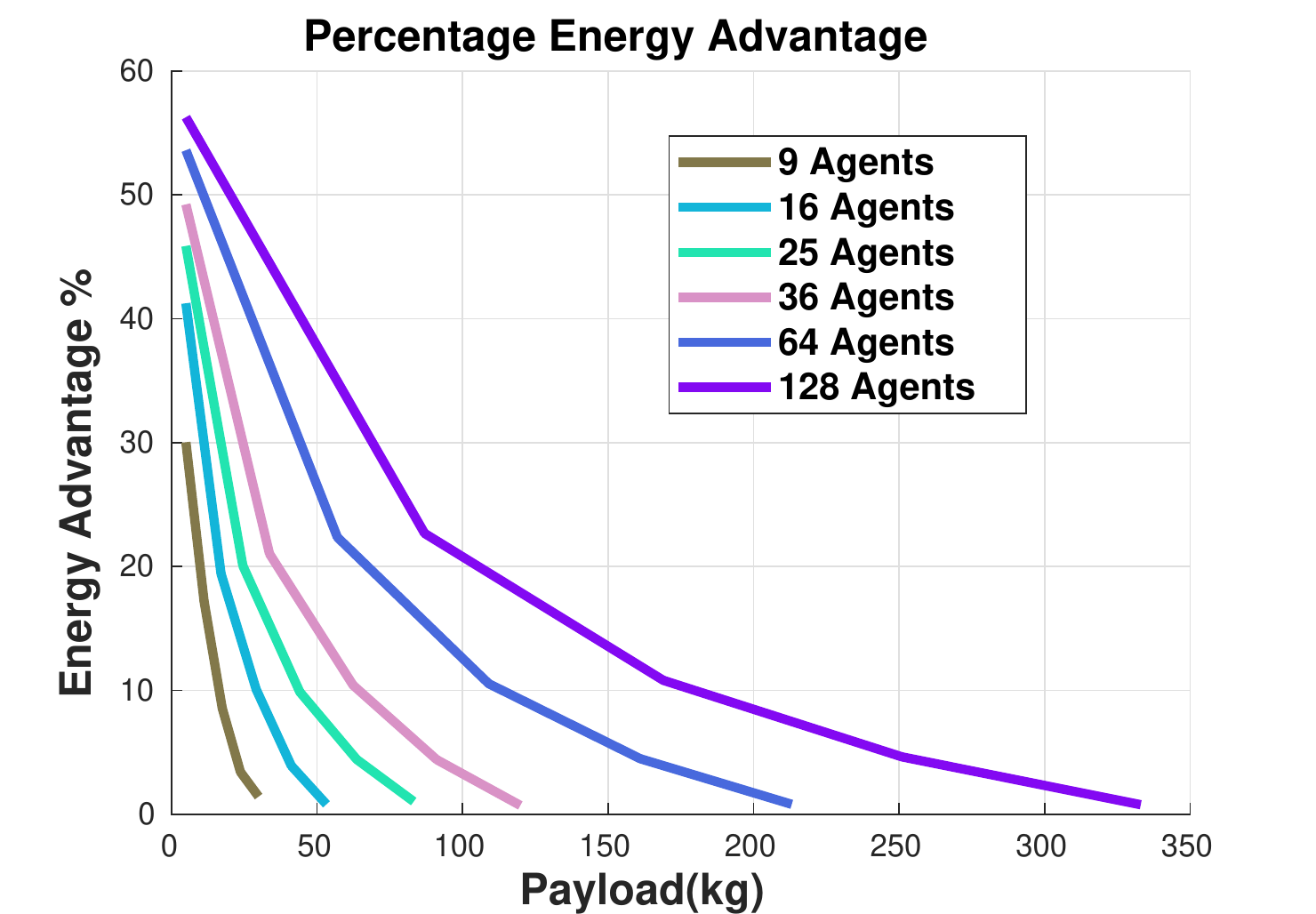}	
	\caption{Energy Advantage  v/s Number of agents}
	\label{eff_result_3}
\end{figure}

\underline{Up-Scaling Number of Agents:} Increasing the number of agents increases the load carrying capacity of the system. Moreover, we observed that by using our energy conscious algorithm (\ref{Efficiency}) the percentage of power saving improves, when total number of agents in the system is increased, as shown in Fig. \ref{eff_result_3}. We also observe that the energy advantage curve has reduced slope with agent scale up, which indicates that the advantage is preserved for a larger range of payload weights.


\begin{figure}
\centering
\includegraphics[width = \columnwidth]{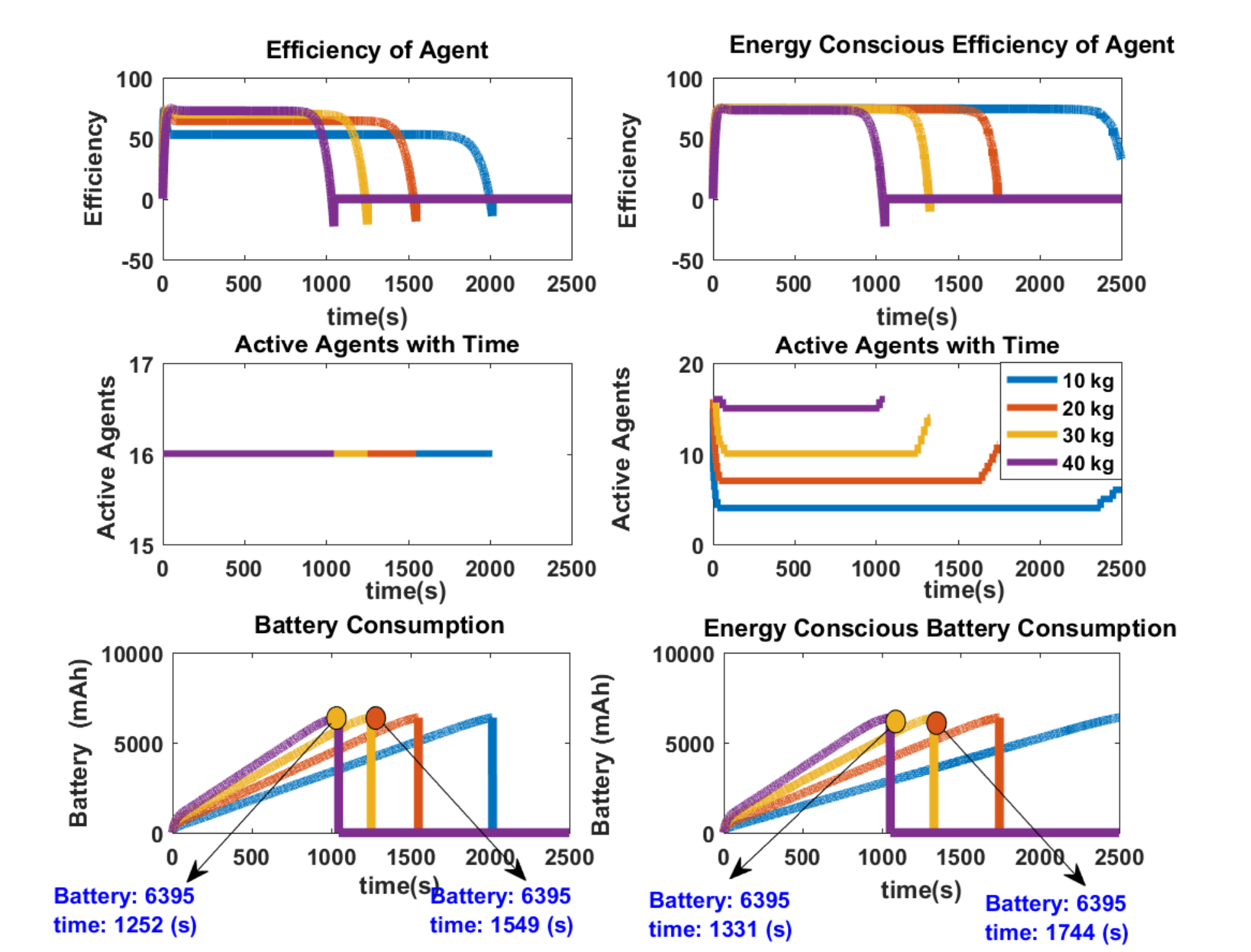}
\caption{ Comparison of battery discharge behaviors for non-energy conscious and energy conscious operations. Notice that the system operates for a longer duration with energy conscious operation.}
\label{bateff}
\end{figure}

\underline{Effect of Battery Discharge:} Fig. \ref{bateff} compares straight-line energy conscious and non-energy conscious operation for type-I motors with the effect of battery discharge. Coefficients for the two-term exponential model fit (Sec. \ref{sec:41}) are $ h =  -1.851\times 10^{-14}, w =    0.005345, y =  23.4,  z =  -1.018 \times 10^{-5}$.  
Battery discharge reduces the available supply. This affects the number of active agents chosen by the energy conscious operation. { In Fig. \ref{bateff} the first row of subplots compares efficiency of agent v/s time.  Notice that the efficiency of an energy conscious agent is on average higher than a regular agent. In the second row, observe that the number of active agents are optimized in accordance with the payload weight. Towards the end of operation the number of active agents increase for energy conscious operation. This is due to the decrease in supply voltage with battery discharge. Finally, in row three we observe that, in general, the system lasts longer with energy conscious operation. For example, notice the highlighted data points in Fig. \ref{bateff}. For a battery capacity of $6395$ mAh and a payload weight of $30$ kg the system lasts for ($\approx 800s$) longer. }

\begin{figure}
	\centering
	\includegraphics[width = 9.5cm]{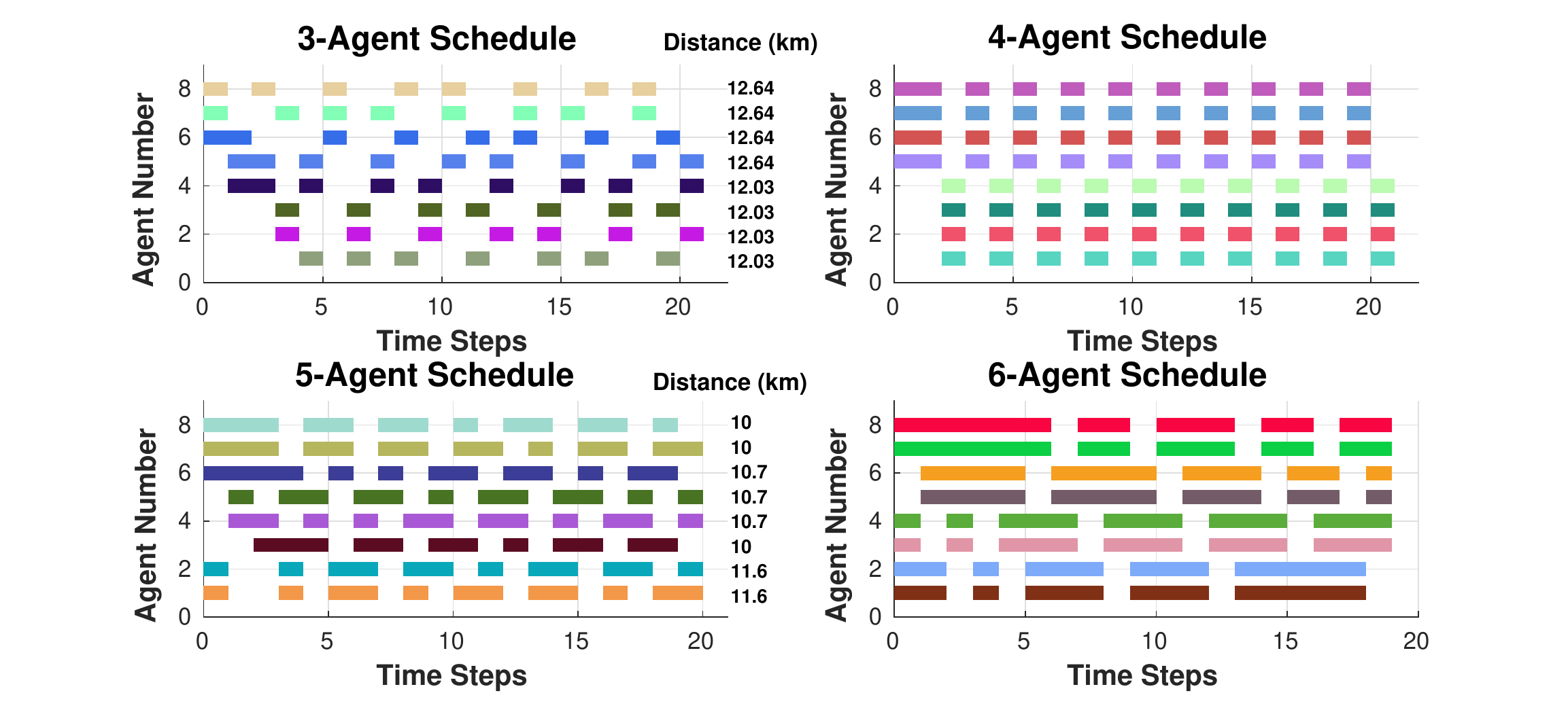}
	\caption{Online: Configuration transitions (1 step=100 s) }
	\label{fig:online}
\end{figure}

\begin{figure}
		\centering
		\includegraphics[width=\columnwidth]{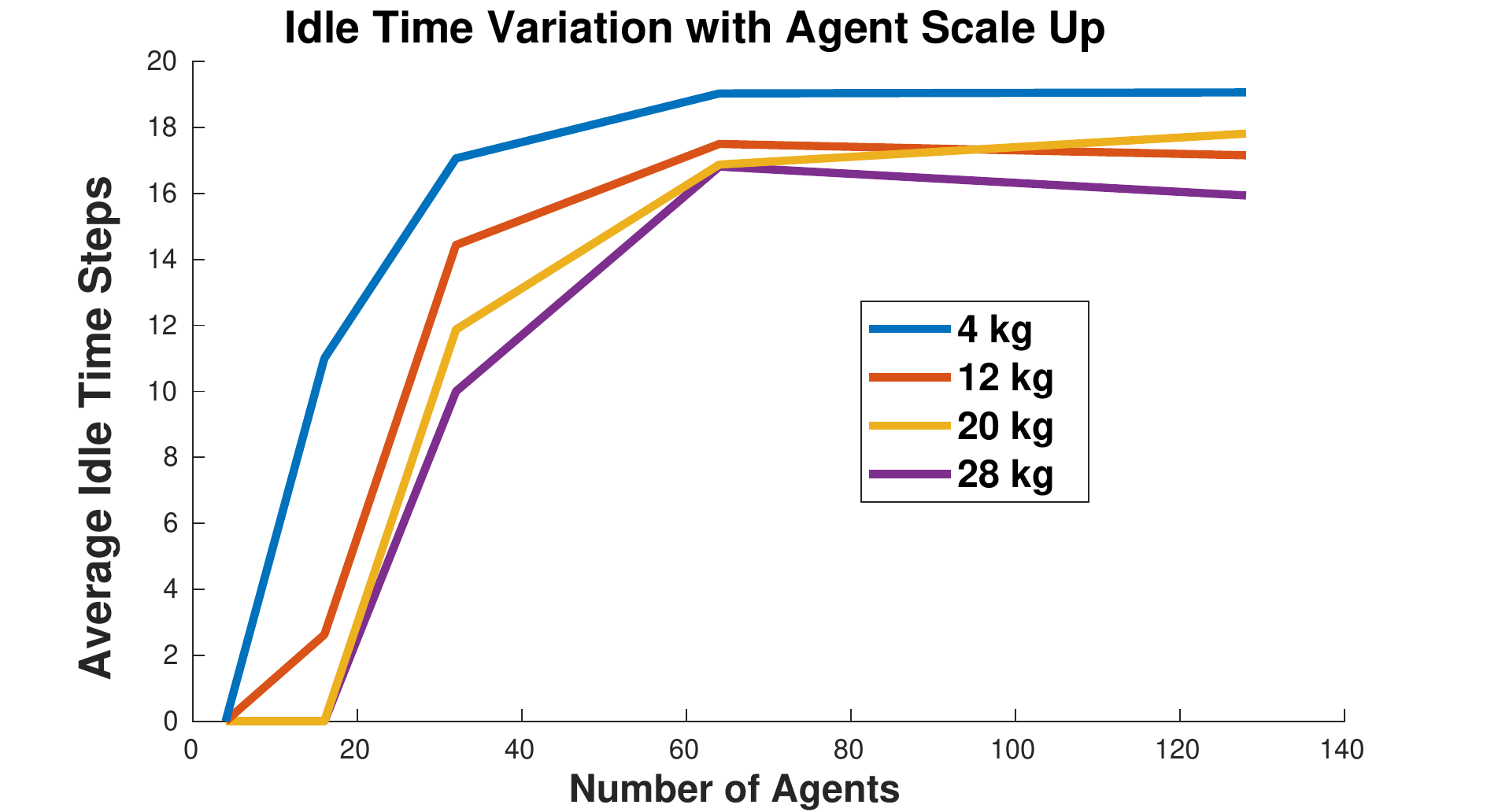}
	\caption{Average agent idle time v/s total number of agents, determined using the proposed multi-level controller.}
	\label{fig:idle}			
\end{figure}

\textbf{Results of Task Allocation:} From basic thermal analysis we observe that the Type-I,II motors should not approximately exceed $600s$ of continuous operation at $1$ Ampere to avoid overheating.  Each time step is $T_{ON} = 100s$. Fig. \ref{fig:online} showcases the results of performing online task allocation, with energy conscious operation, for different number of active agents. Additionally, simulating the three-level control architecture with online task allocation for $30$ minutes, we plot the average idle time steps per agent as the number of agents is scaled up for a known payload weight. Fig. \ref{fig:idle} shows that the average idle time (steps) per agent increases with agent scale up for a constant payload weight. This is an advantage as the agents ensure efficient system operation by staying idle for longer periods of time, resulting in lesser overall motor wear. 

\begin{figure}[h]
	\vspace{-1em}
	\centering
	\includegraphics[width = \columnwidth]{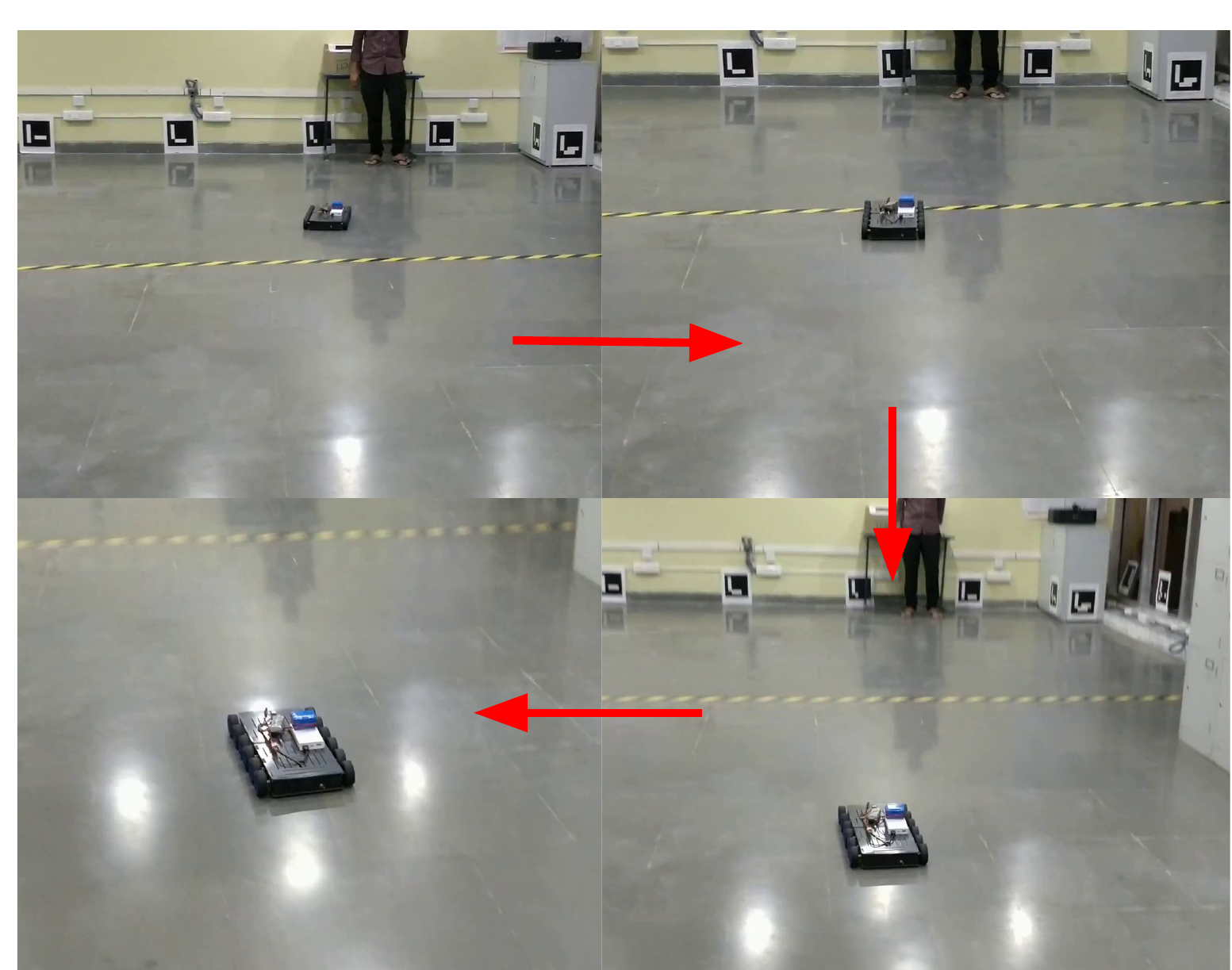}
	\caption{ Experimental Platform in Motion}
	\label{fig:exp1}
\end{figure}

\textbf{Validation of Physical System:} 
We validate (a) the feasibility of using a collection of low-torque motors to transport payloads and (b) de-activating some motors after a few seconds of operation and ensure system motion. Fig. \ref{fig:exp1} showcases a constructed physical model in motion. The physical model consists of $12$ low torque and free moving Type-III motors (Table \ref{tab:1}). We use an Arduino Mega Central Controller in conjunction with a Raspberry Pi 3 micro-computer running ROS \cite{quigley2009ros}. The system costs under $400 \$$, weighs $3 kg$ and can carry a payload weight of upto $10$ kg (Type-III motors).

%% file: conclusions.tex
\section{Conclusion} \label{sec:7}
\noindent In this work, we present a novel energy conscious over-actuated robot which selectively actuates only a subset of all its available wheels. Decisions on how many and which of the wheels to activate are made in real-time using a hierarchical decision making architecture. Energy conscious operation (Sec. \ref{sec:52}) is achieved by leveraging non-linear optimization. 
Furthermore, the system ensures that all its agents are utilized uniformly to minimize agent wear (Sec. \ref{sec:54}). Our system can energy consciously track any trajectory, using the developed kinematic, dynamic and electric models (Sec. \ref{sec:3}). { Further work would involve introduction of holonomic, energy conscious steering for our over-actuated system. This would aid further development of modular and decentralized multi-agent controllers. Our system and the developed control methodologies could enable development of future human transport systems.}